\documentclass{svproc}
\usepackage{amsmath}
\usepackage{booktabs}
\usepackage{url}

\usepackage{graphicx}
\usepackage{multirow}

\begin{document}
\mainmatter 

% -----
\title{Multimodal Deep Learning for Subtype Classification in Breast Cancer Using Histopathological Images and Gene Expression Data}
\titlerunning{Multi-Modal Fusion for Breast Cancer Prediction} 

\author{Dr. Amin Honarmandi Shandiz}
\authorrunning{Dr. A. H. Shandiz} 

\tocauthor{A. H. Shandiz et al.} 

\institute{Szeged, Hungary\\
\email{Amin.HonarmandiShandiz@gmail.com}
}

\maketitle  

\begin{abstract}
Molecular subtyping of breast cancer is crucial for personalized treatment and prognosis. Traditional classification approaches rely on either histopathological images or gene expression profiling, limiting their predictive power. In this study, we propose a deep multimodal learning framework that integrates histopathological images and gene expression data to classify breast cancer into BRCA.Luminal and BRCA.Basal / Her2 subtypes. Our approach employs a ResNet-50 model for image feature extraction and fully connected layers for gene expression processing, with a cross-attention fusion mechanism to enhance modality interaction. We conduct extensive experiments using five-fold cross-validation, demonstrating that our multimodal integration outperforms unimodal approaches in terms of classification accuracy, precision-recall AUC, and F1-score. Our findings highlight the potential of deep learning for robust and interpretable breast cancer subtype classification, paving the way for improved clinical decision-making.\\
The code for our proposed multimodal deep learning framework is available at \url{https://github.com/AminHonarmandiShandiz/cancerpredict}.

\keywords{Breast Cancer Subtyping, Multimodal Deep Learning, Histopathological Image Analysis, Gene Expression Profiling, Cross-Attention Fusion, Convolutional Neural Networks (CNN), Molecular Classification, Precision Oncology, Feature Alignment, Cancer Heterogeneity.}
\end{abstract}
\section{Introduction}
Breast cancer is a heterogeneous disease, and precise molecular subtyping is essential for effective treatment selection and prognosis prediction~\cite{parker2009supervised}. Traditional classification approaches, including histopathological evaluation and genomic profiling, have been widely used in clinical practice~\cite{young2016pathway,spanhol2016breast}. However, single-modality methods often fail to capture the full complexity of tumor biology. While histopathological images provide morphological and structural insights, gene expression data offer a detailed molecular landscape, including information on oncogenic pathways and tumor microenvironment interactions~\cite{krawczuk2016feature,jass2007classification}. Despite their individual strengths, these modalities have limitations when used independently, leading to suboptimal classification performance and reduced clinical applicability.

Recent advancements in deep learning have demonstrated promising results in cancer classification by leveraging convolutional neural networks (CNNs)~\cite{spanhol2016breast} for imaging data and deep neural networks for genomic data~\cite{chen2016gene}. While there are several multi modal setup using AI in different applications such as Speech and ultrasound~\cite{yu2021reconstructing,toth20203d,csapo2022optimizing,shandiz2021neural,shandiz2021improving,honarmandi2021voice,toth2023adaptation,zainko2021adaptation,shandiz2022improved} Inspired by the DeepCC framework~\cite{gao2019deepcc}, which transformed gene expression profiles into functional spectra for cancer subtype classification, we propose a multimodal deep learning approach that integrates both histopathological images and gene expression data to improve breast cancer subtyping. This study explores the effectiveness of different fusion strategies, including cross-attention fusion, concatenation fusion, and late fusion, to determine the optimal method for integrating multimodal information. By enhancing feature alignment and leveraging complementary information from both data sources, our framework aims to improve the accuracy and robustness of breast cancer subtype classification.

\section{Materials and Methods}
\subsection{Dataset and Pre-processing}
\begin{figure}[ht]
\includegraphics[width=\textwidth]{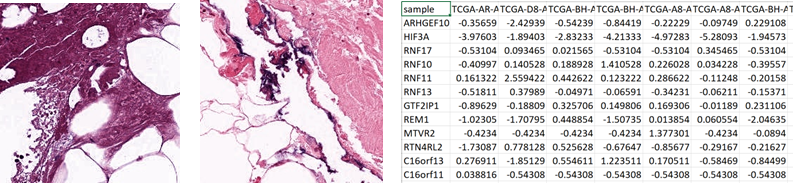}
\caption{An example of Histopathology images(left) and Gen expression(right) samples.} \label{fig:data}
\end{figure}

This study utilizes a dataset containing histopathological images and gene expression profiles from breast cancer patients shows in Figure~\ref{fig:data}~\cite{dataset_2025}. 
The histopathological images were processed by extracting 256×256 pixel patches from whole-slide images, ensuring that only high-content tissue regions were selected using Otsu’s thresholding. 

\begin{figure}[ht]
\includegraphics[width=\textwidth]{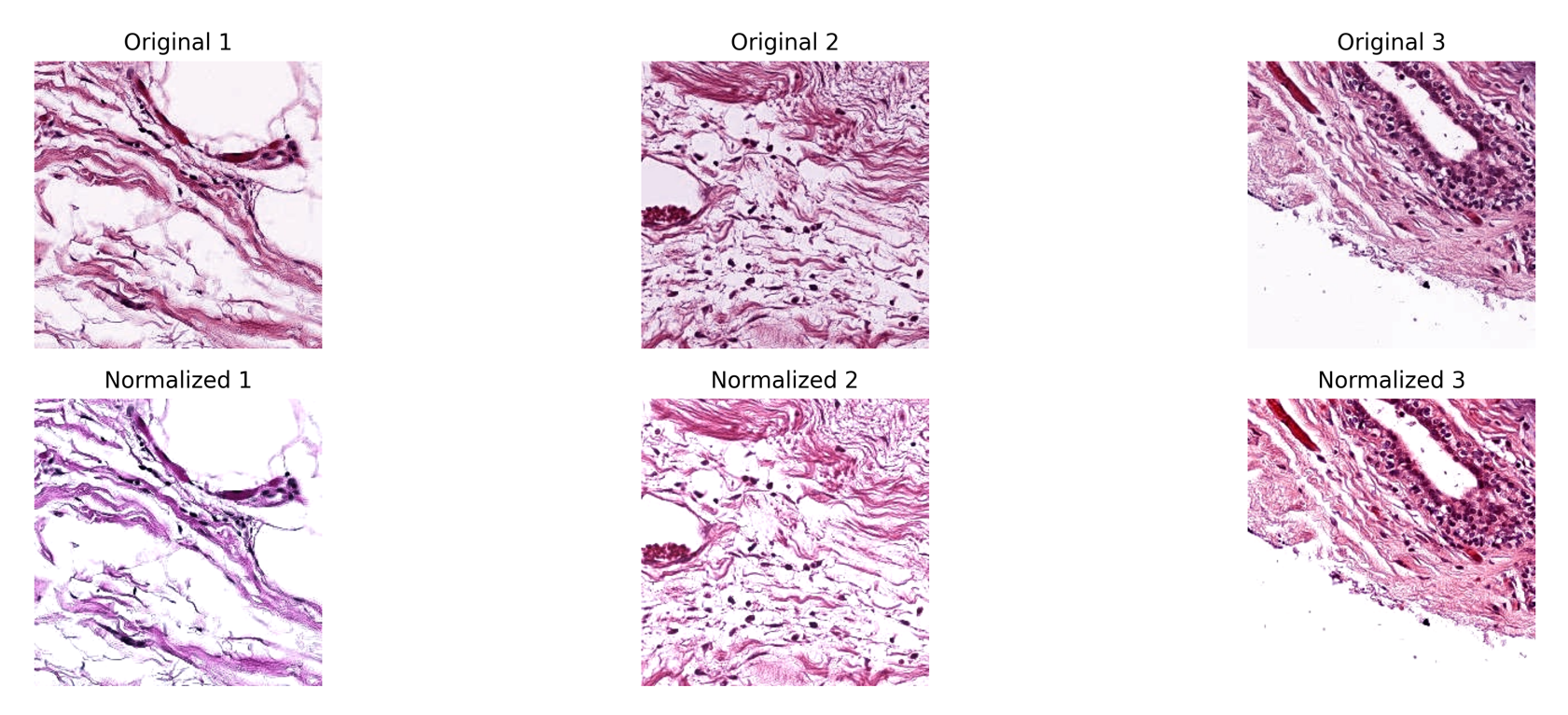}
\caption{Figure illustrate before (up) and after (down) image augmentation of selected patches.} \label{fig:histpreprocessing}
\end{figure}

To enhance model generalization, data augmentation techniques such as color jitter with Brightness 20\%, Contrast 20\%, Saturation 20\% and Hue 10\% and random rotations were applied to the image patches as Figure~\ref{fig:histpreprocessing} shows. Feature extraction was performed using a ResNet-50 model pre-trained on ImageNet, producing a 35×2048 feature matrix per patient.

\begin{figure}[!h]
\includegraphics[width=\textwidth]{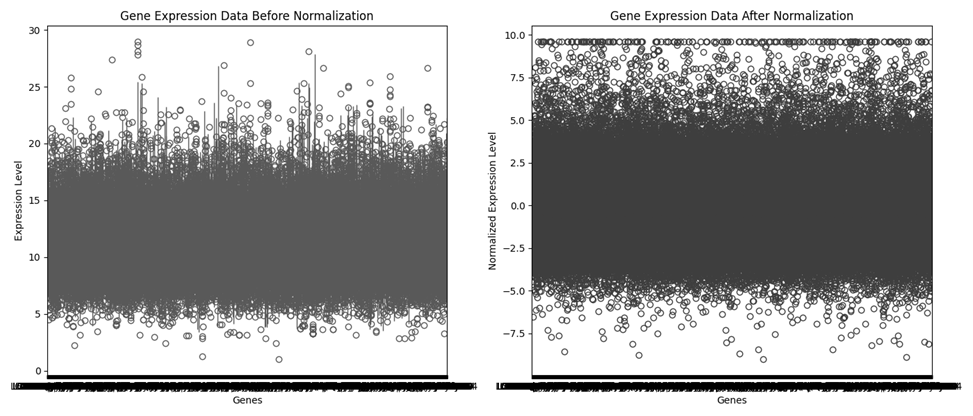}
\caption{The Figure shows the Gen expression before(left) and after(right) z-score normalization.} \label{fig:genprocessing2}
\end{figure}

For the gene expression data, preprocessing involved multiple steps to ensure consistency and accuracy. Negative values were adjusted by shifting all values by the absolute minimum plus one, followed by log transformation and Z-score normalization shown in Figure~\ref{fig:genprocessing2}.

\begin{figure}[ht]
\includegraphics[width=\textwidth]{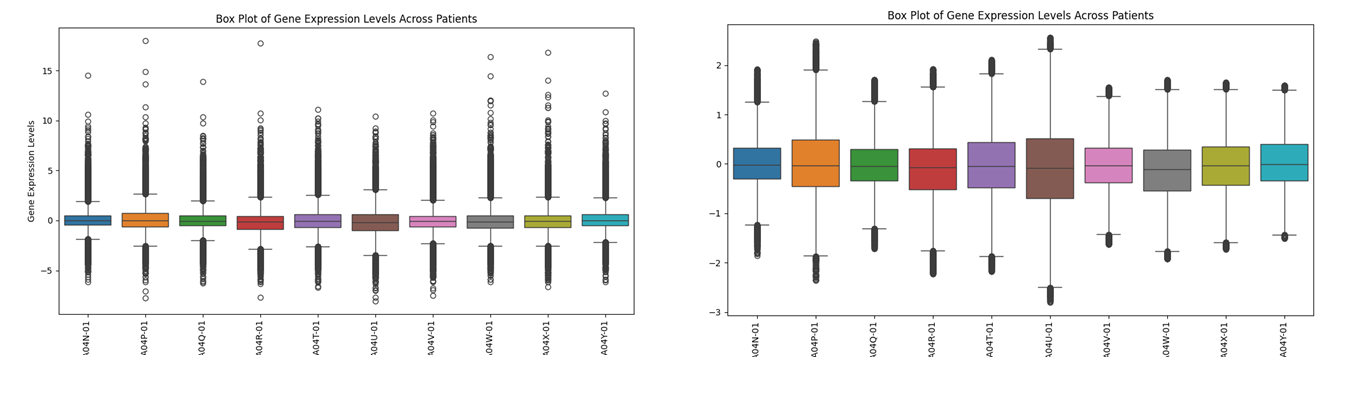}
\caption{Ten patient gen expression plot before and after outlier detection.} \label{fig:genprocessing}
\end{figure}

Outliers were identified and removed using the interquartile range (IQR) method in Figure~\ref{fig:genprocessing}, and missing values were imputed using either the mean or median strategy. To align samples across modalities, gene expression data were filtered to include only patients with corresponding histopathological images. Class labels were assigned based on molecular subtyping, with BRCA.Luminal labeled as 0 and BRCA.Basal/Her2 labeled as 1.

\subsection{Model Implementation}
\begin{figure}[!h]
\includegraphics[width=\textwidth]{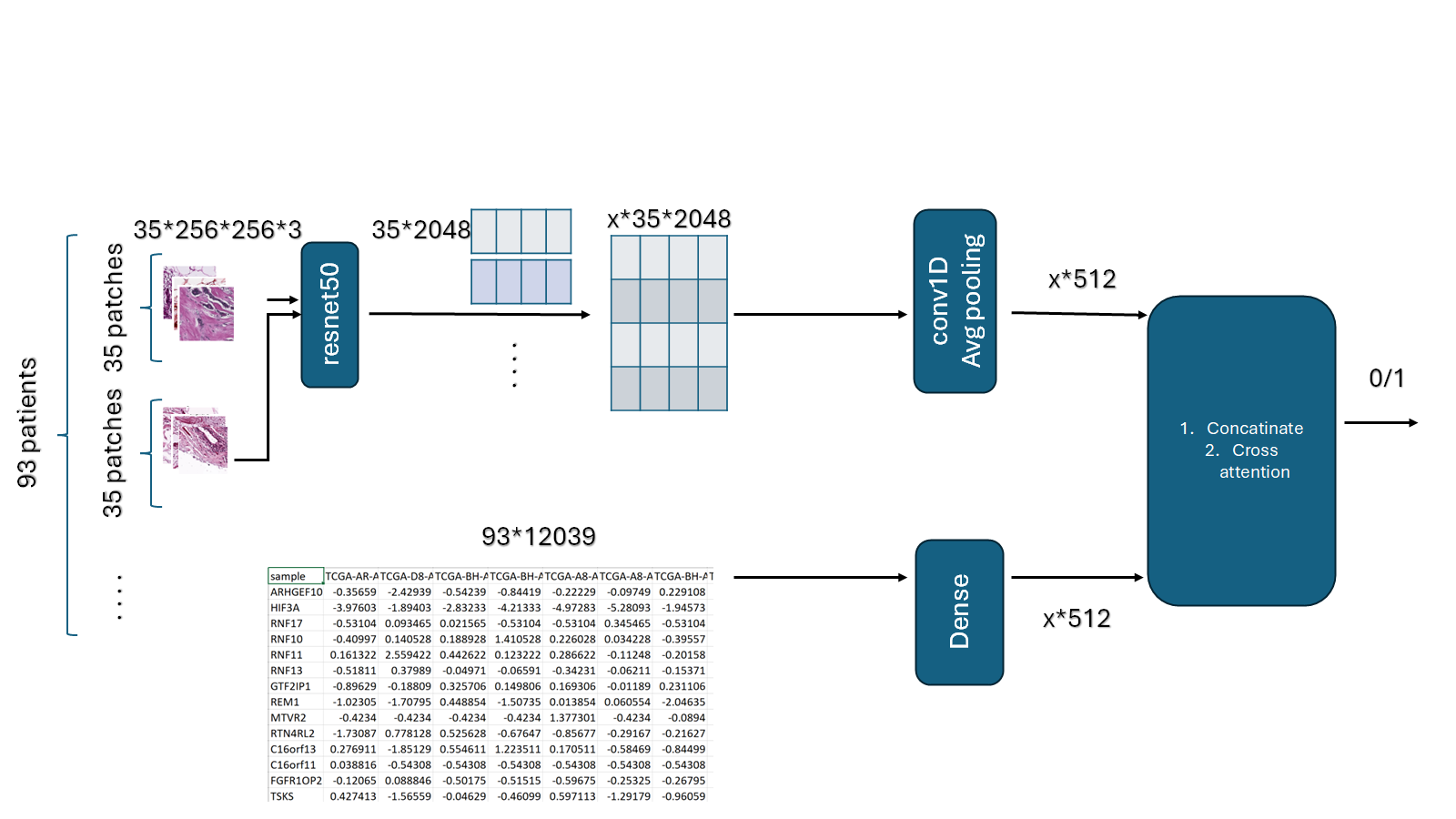}
\caption{The Figure shows the multi-modal fusion setup.} \label{fig:model}
\end{figure}

The deep learning model consists of two parallel pathways for feature extraction, followed by a fusion mechanism, in figure~\ref{fig:model}. The histopathology pathway processes image features using ResNet-50, extracting a compact feature representation. The genomic pathway utilizes a fully connected neural network to encode gene expression profiles into a latent space. Three fusion strategies were evaluated: concatenation fusion, where features from both modalities are merged directly; late fusion, where predictions from separate unimodal models are combined; and cross-attention fusion, where a multi-head attention mechanism is applied to align features from histopathological images and gene expression data. The cross-attention mechanism was designed to enhance feature interactions by dynamically weighting the importance of each modality’s features.

Model training was conducted using the Adam optimizer with a learning rate of 1e-5, and binary cross-entropy was used as the loss function. Five-fold cross-validation was implemented to ensure robust performance evaluation, and performance metrics included F1-score, MCC, and PR-AUC to assess classification effectiveness.

\section{Results}
\subsection{Unimodal Performance}

\begin{figure}[!h]
\includegraphics[width=\textwidth]{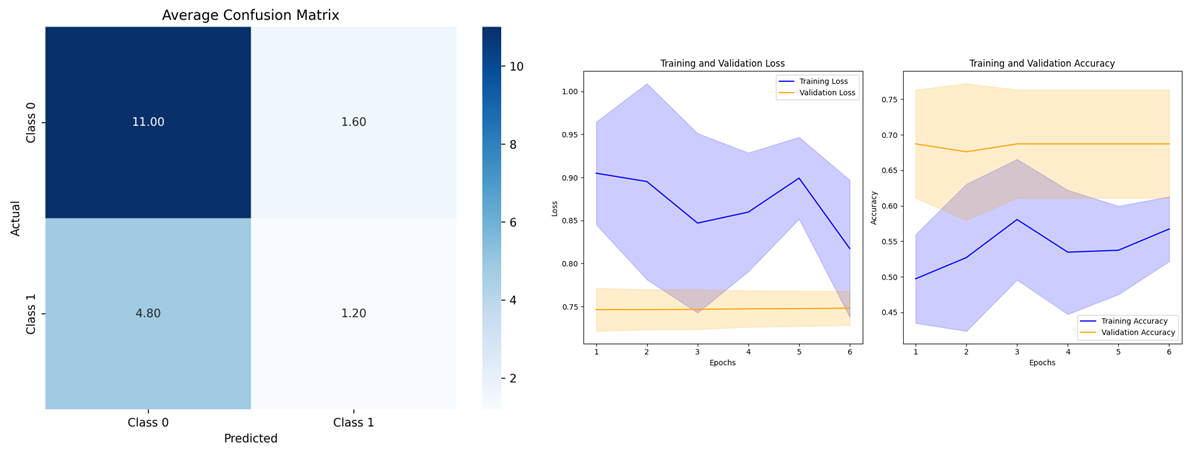}
\caption{The Figure shows the 5-fold cross validation of the confusion matrix and training curve  for single image modality.} \label{fig:just-image}
\end{figure}

\begin{figure}[!h]
\includegraphics[width=\textwidth]{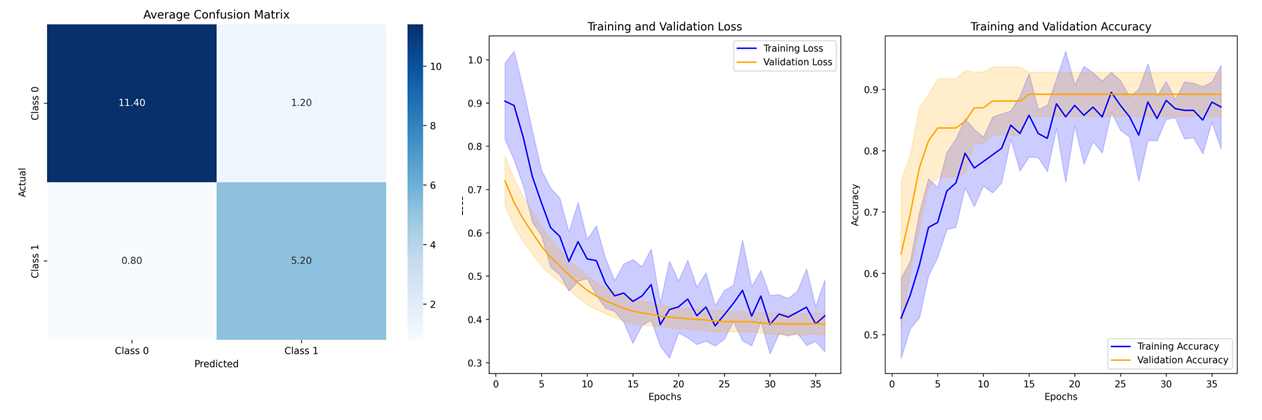}
\caption{The Figure shows the evaluation result of the model with Gen data as input.} \label{fig:just-gen}
\end{figure}

Initial experiments assessed the performance of single-modality models. The gene expression model achieved an F1-score of 0.8197 and a PR-AUC of 0.9435, demonstrating strong predictive power from genomic data alone. However, the histopathological image model underperformed, with an F1-score of 0.1780 and a PR-AUC of 0.3656 O(Figure~\ref{fig:just-image} and Figure~\ref{fig:just-gen} respectively), indicating that image features alone were insufficient for accurate subtype classification. While training curve of the image modality shows even low precise as the trainings has ran with an early stopping call-back therefore the curve is average with the minimum cut steps over 5-Fold.

\subsection{Multimodal Performance}
\begin{figure}[!h]
\includegraphics[width=\textwidth]{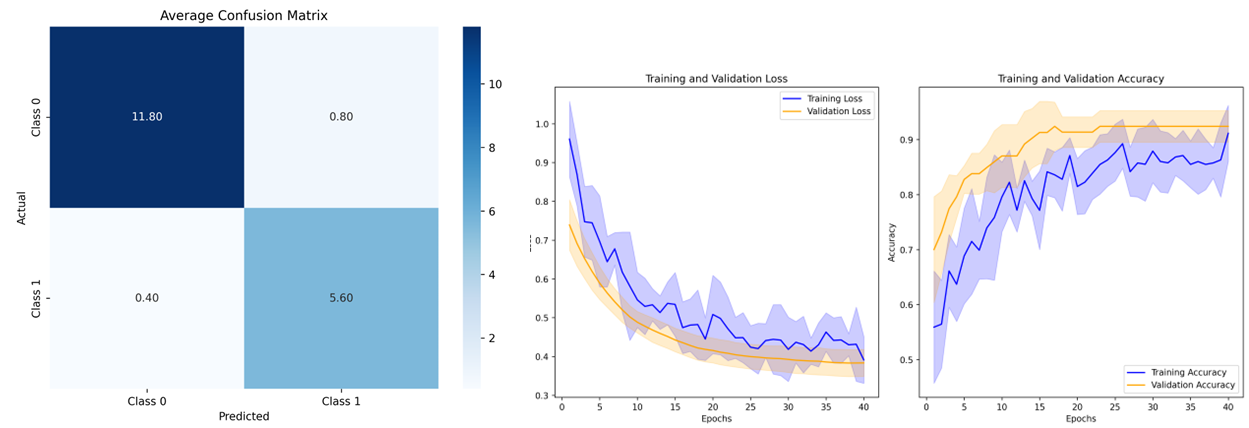}
\caption{The Figure shows the multi modal results with concatination fusion.} \label{fig:multi-fusion}
\end{figure}

\begin{figure}[!h]
\includegraphics[width=\textwidth]{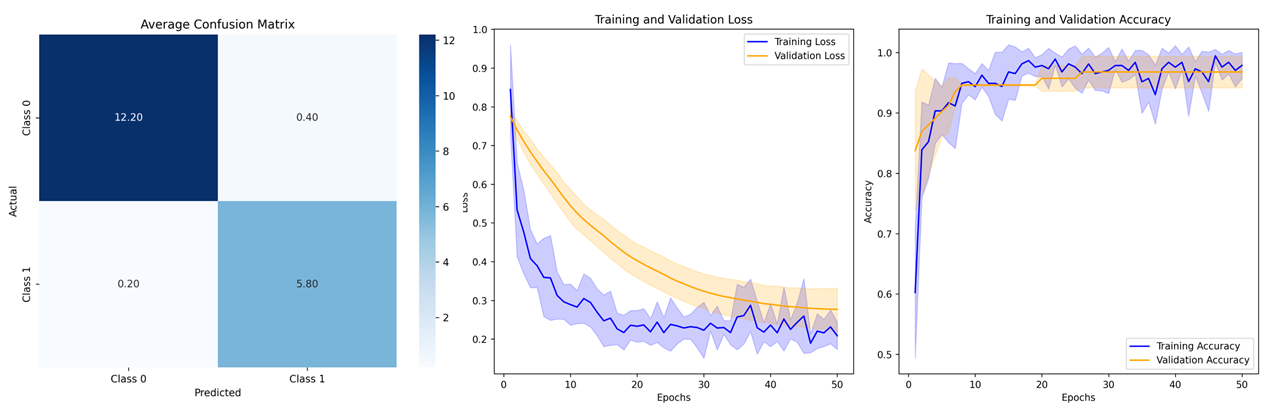}
\caption{The Figure shows The multi-modality result with cross-attention fusion.} \label{fig:multi-attention}
\end{figure}

The integration of both modalities significantly improved classification performance. The cross-attention fusion model achieved the highest accuracy, with an F1-score of 0.9379 and a PR-AUC of 0.9948, Figure~\ref{fig:multi-attention}. The concatenation fusion model followed closely, with an F1-score of 0.8960 and a PR-AUC of 0.9684 (figure~\ref{fig:multi-fusion}), while the late fusion approach yielded lower performance, confirming that feature-level integration is more effective than decision-level fusion.

\subsection{Ablation Study}
To evaluate the contribution of each modality, ablation studies were conducted. Removing image features led to a 12\% drop in classification accuracy, confirming the added value of histopathology data when combined with gene expression. Eliminating the cross-attention mechanism resulted in a 5\% reduction in PR-AUC, demonstrating the effectiveness of attention-based feature alignment.

\subsection{Confusion Matrix Analysis}
Analysis of the confusion matrix showed high recall for BRCA.Luminal cases, while BRCA.Basal/Her2 cases exhibited moderate misclassification, suggesting further optimization may be required for certain subtypes.

\section{Discussion}
The results highlight the advantages of multimodal deep learning in breast cancer classification. The integration of histopathological and genomic features significantly enhances predictive performance compared to unimodal approaches. The superior performance of the cross-attention fusion model underscores the importance of aligning and weighting complementary features from different modalities.

Clinically, this approach has the potential to improve patient stratification and treatment selection by providing more accurate molecular subtype predictions. The study also demonstrates the importance of preprocessing steps, particularly outlier removal and normalization, in stabilizing model performance. However, some limitations remain, including dataset size constraints and the need for external validation on independent cohorts. Future work will explore transformer-based fusion models and interpretability techniques to enhance clinical applicability.

\section{Conclusion}
This study presents a novel deep multimodal learning framework for breast cancer subtype classification. By integrating histopathological images and gene expression data using a cross-attention fusion model, the proposed approach achieves state-of-the-art performance in classification accuracy and robustness. These findings highlight the potential of deep learning to advance precision oncology by providing more reliable and interpretable cancer subtyping solutions.
\newpage
\bibliographystyle{IEEEtran}

\bibliography{CIARP2021.bib}

\end{document}